\documentclass[10pt,twocolumn,letterpaper]{article}

\usepackage{ijcb}
\usepackage{times}
\usepackage{epsfig}
\usepackage{graphicx}
\usepackage{amsmath}
\usepackage{amssymb}
\usepackage[norule,symbol,perpage]{footmisc}
\usepackage{url}

\ijcbfinalcopy 

\ifijcbfinal\fi

\makeatletter
\def\ps@IEEEtitlepagestyle{
\def\@oddfoot{\mycopyrightnotice}
\def\@evenfoot{}
}
\def\mycopyrightnotice{
{\hfill \footnotesize 978-1-7281-9186-7/20/\$31.00 \copyright 2020 IEEE\hfill}
}
\makeatother

\begin{document}

\title{How Do the Hearts of Deep Fakes Beat? \\ Deep Fake Source Detection via Interpreting Residuals with Biological Signals}

\author{Umur Aybars Ciftci\\
Binghamton University\\
{\tt\small uciftci@binghamton.edu}
\and
\.{I}lke Demir\\
Intel Corporation\\
{\tt\small  idemir@purdue.edu}
\and
Lijun Yin\\
Binghamton University\\
{\tt\small  lijun@cs.binghamton.edu}
}

\maketitle
\thispagestyle{empty}

\begin{abstract}
Fake portrait video generation techniques have been posing a new threat to the society with photorealistic deep fakes for political propaganda, celebrity imitation, forged evidences, and other identity related manipulations. Following these generation techniques, some detection approaches have also been proved useful due to their high classification accuracy. Nevertheless, almost no effort was spent to track down the source of deep fakes. We propose an approach not only to separate deep fakes from real videos, but also to discover the specific generative model behind a deep fake. Some pure deep learning based approaches try to classify deep fakes using CNNs where they actually learn the residuals of the generator. We believe that these residuals contain more information and we can reveal these manipulation artifacts by disentangling them with biological signals. Our key observation yields that the spatiotemporal patterns in biological signals can be conceived as a representative projection of residuals. To justify this observation, we extract PPG cells from real and fake videos and feed these to a state-of-the-art classification network for detecting the generative model per video. Our results indicate that our approach can detect fake videos with 97.29\% accuracy, and the source model with 93.39\% accuracy. 
\end{abstract}

\let\thefootnote\relax\footnotetext{\mycopyrightnotice}

\section{Introduction}

\begin{figure*}[ht]
\centering
    \includegraphics[width=0.95\textwidth]{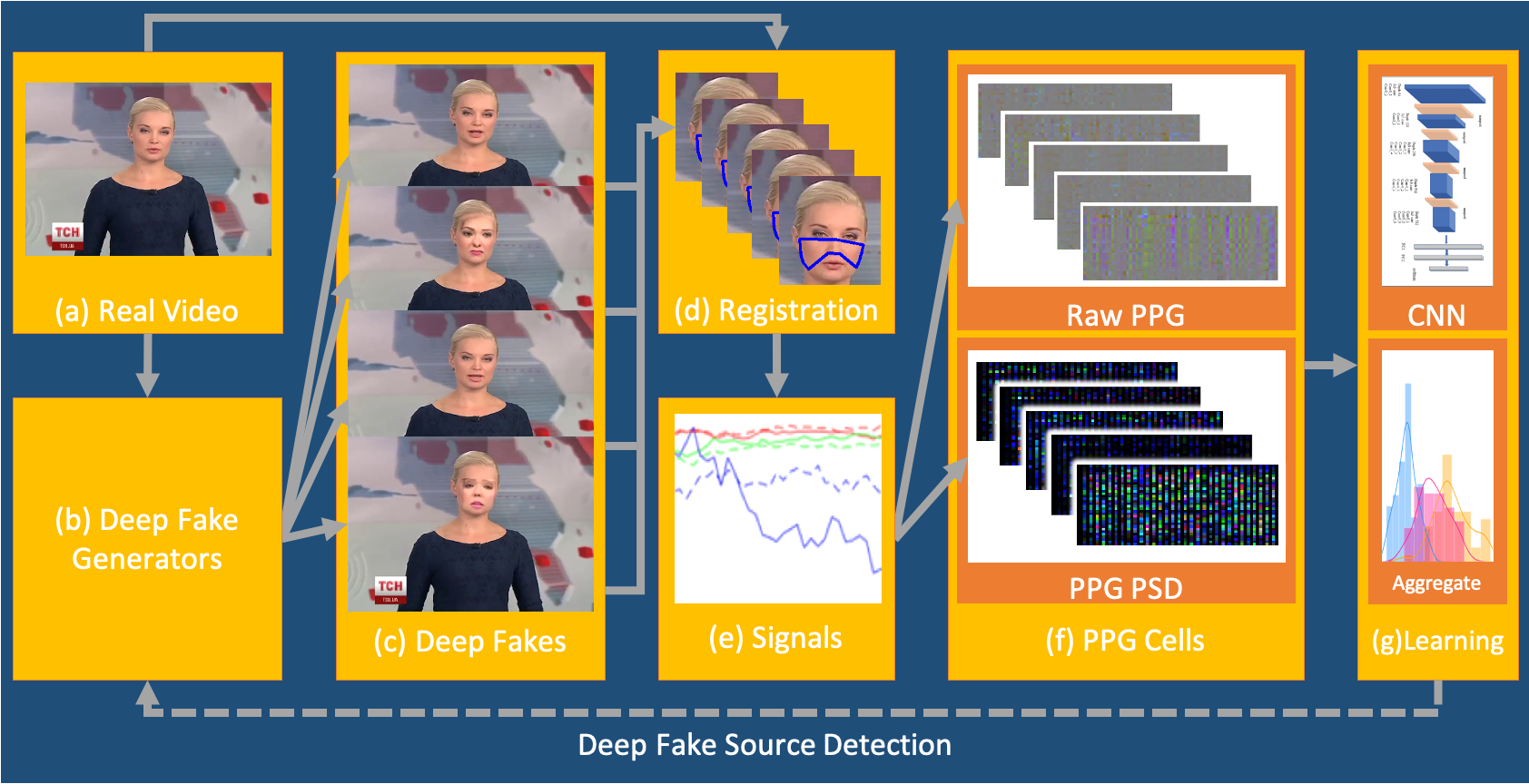}
    \caption{\textbf{Overview.} From real videos (a), several generators (b) create deep fakes with residuals specific to each model (c). Our system extracts face ROIs (d) and biological signals (e), to create PPG cells (f) where the residuals are reflected in spatial and frequency domains. Then it classifies both the authenticity and the source of any video (c) by training on PPG cells and aggregating window predictions (g).}
  \label{fig:teaser}
  \end{figure*}
  
Artificial intelligence (AI) approaches to generate synthetic videos~\cite{10.1145/3306346.3323035, Karras_2019_CVPR, DBLP:journals/corr/abs-1710-10196} have reduced required level of skill for realistic image manipulation~\cite{Wang_2019_ICCV, FaceSwap}. These advancements precipitated the rise of deep fakes \cite{DeepFakes, FakeApp}, synthetic portrait videos of real humans, photorealistic enough to be used as fakes. Although this technology has been developed with positive intent for movies~\cite{DeepFakeDub, DeepFakeMovies}, advertisement~\cite{DeepFakeAds}, virtual clothing ~\cite{6588873}, and entertainment~\cite{DeepFakeEnt}; unfortunately this strong impact attracted malicious users to exploit deep fakes for political misinformation~\cite{DeepFakePol} and pornography~\cite{DeepFakePorn}. 

This threat to information integrity has consequences in privacy, law, politics, security, and policy, and has the potential to form a social erosion of trust~\cite{ucla}. As a defense mechanism, deep fake detection methods have been introduced~\cite{9065881, 8744516, blink}, which define the problem as a binary classification. It is conceivable that, as more realistic and complex generation methods are developed over time, detection methods should also have a more profound development and a deeper understanding. Deciding the authenticity of a video is demanded, however finding the source is even more important and challenging for tracking, prevention, and combating their spread. We 
propose a deep fake source detector that predicts the source generative model for any given video. To our knowledge our approach is the first to conduct a deeper analysis for source detection that interprets residuals of generative models for deep fake videos.

Biological signals are present in all humans. Anatomical actions such as heart beat, blood flow, or breathing, create subtle changes that are not visible to the eye but still detectable computationally. For example, when blood moves through the veins, it changes the skin reflectance over time, due to the hemoglobin content in the blood. Approaches to extract photoplethysmography (PPG) signals are developed to recognize such changes by image processing techniques. 

As of now, no generative model is able to create deep fakes with consistent PPG signals. Several previous approaches utilize similar biological signals to detect 3D synthetic CG renders~\cite{7025049} and deep fakes~\cite{FakeCatcher, blink, 8683164}. \cite{FakeCatcher} in particular proves that spatiotemporal inconsistency of biological signals can be exploited to detect deep fakes. Our key observation follows the fact that biological signals are not yet preserved in deep fakes, and those signals produce different signatures in terms of the generative noise. Thus, we can interpret biological signals as a projection of the residuals in a known dimension that we can explore to find the unique signature per model. This motivates us to utilize these signals for the recognition of the current and future generative models behind all deep fake videos. More importantly, the source detection can also improve the overall fake detection accuracy, because real videos with inconsistent biological signals (due to occlusions, illumination changes, etc.) can still be recognized as real videos as they do not conform to the signature of any generative model. 

In our work, we extract 32 raw PPG signals from different locations in the face, from a window of frames, from a video of windows. We then encode the signals along with their spectral density into a spatiotemporal block, which is so-called \textit{PPG cell}. We feed PPG cells into an off-the-shelf neural network to recognise the signatures of the distinct residuals of the source generative models. Lastly, we combine per sequence predictions into a per video prediction using average log of odds~\cite{Hsiao1996}. Our approach achieves the prediction for the authenticity of the video by 97.29\%, and the generative model by 93.39\% on FaceForensics++~\cite{FF++} dataset. We evaluate our approach on five datasets with multiple~\cite{FF++}, single~\cite{Celeb_DF_cvpr20, 8683164, ff}, and unknown~\cite{FakeCatcher} generators, against five state of the art source models, and seven backbones. We also conduct an ablation study on various setups for comparison, and compel our approach towards extension to new models and detecting unseen generators.
 
In summary, the contributions of this paper are listed as,
\vspace{-0.5em}
\begin{itemize}
  \setlength\itemsep{-0.5em}
  \item a novel approach for deep fake source detection, leading deep fake detection research to a new perspective, 
  \item a new discovery that the projection of generative noise into biological signal space can create unique signatures per model, and
  \item an advanced general deep fake detector that can outperform current approaches in fake/real classification, while also predicting the source generative model. 
\end{itemize}

\section{Related Work}

\subsection{Generative Models for Deep Fakes}

There exist various deep fake methods in the literature \cite{DeepFakes,f2f,10.1145/3306346.3323035,10.1145/2816795.2818056,FakeApp,FaceSwap,Karras_2019_CVPR,9065881}. We categorize these methods broadly based on their face synthesis as (i) face generation, (ii) face reenactment, and (iii) face manipulation techniques. The first category for face generation mostly consists of generative adversarial network (GAN) based methods. For example, StyleGAN~\cite{Karras_2019_CVPR} and ProGAN~\cite{DBLP:journals/corr/abs-1710-10196} are methods to create entire fake faces. The second category for face reenactment includes the face replacement methods using model warping or swapping techniques, for example using a 3D model of another person such as ~\cite{DeepFakes,f2f,10.1145/3306346.3323035,FaceSwap,8373817,10.1145/2816795.2818056,Garrido_2014_CVPR,Wu_2018_ECCV}. The third category for face manipulation mostly focuses on facial expression transfer or mouth shape and movement synthesis from lip reading, while keeping the face identity intact~\cite{10.1145/2816795.2818056}. 

\subsection{Deep Fake Detectors}

For fake image detection from the face generation category, several typical signatures have been identified including saturation cues \cite{8803661}, frequencies of generated images for fingerprints of GAN models \cite{Yu_2019_ICCV}, and discrete cosine transformation residuals \cite{dctforgery}.  

For the facial reenactment, detection is usually performed per frame, which also utilizes temporal information. To search for some artifacts which may occur due to the facial differences between the source and target faces, Boulkenafet et al. \cite{boulkenafet16} estimate distortions in the generated faces, Barin et al. \cite{barni17} investigate compression artifacts, Yang et al. \cite{8683164} recognize inconsistent head poses, Li et al. \cite{blink} detect blinking effects, and Li and Lyu \cite{Li_2019_CVPR_Workshops} search for face warping artifacts in the generated faces successfully. In addition, other markers such as biological signals \cite{FakeCatcher} and lighting inconsistency \cite{10.1117/12.2520546} have also been explored. Specific generative networks ~\cite{mesonet,shallownet,8639163,8553251,8124497,8014963,8682602} have been applied as the discriminators. 

Similarly, above methods can be used for detection of face manipulation tasks. Beyond that, motion and extra modality can also be used as auxiliary components to facilitate detection, e.g., inconsistent mouth movements   \cite{8638330} and \cite{8553270}, and audio/visual verification \cite{Korshunov:270130} and \cite{vidTimid}.  

\subsection{Source Detectors}
To our knowledge, most existing deep fake source detectors are image-based, and exploit various attributes of synthetic imagery such as GAN model fingerprints ~\cite{Yu_2019_ICCV,8713484,8695364,Guarnera_2020_CVPR_Workshops}, camera patterns~\cite{1634362}, or image attribution~\cite{Albright_2019_CVPR_Workshops}.
Yu et al. \cite{Yu_2019_ICCV} identify fully synthetic images that are generated with ProGAN~\cite{DBLP:journals/corr/abs-1710-10196}, SNGAN~\cite{DBLP:journals/corr/abs-1802-05957}, CramerGAN~\cite{DBLP:journals/corr/BellemareDDMLHM17}, MMDGAN~\cite{bikowski2018demystifying} and analyze their fingerprints through frequency analysis. 
Lukas et al. \cite{1634362} and Cozzolino et al. \cite{8713484} analyze camera sensor noise for natural images. Marra et al. \cite{8695364} find GAN residual fingerprints in final synthetic image patterns. Albright and McCloskey \cite{Albright_2019_CVPR_Workshops} examine the source camera attribution on GANs. Although all of those approaches can be regarded as an interpretation of the generative noise in a specific GAN generated image domain, they have not been evaluated on deep fake videos yet, neither have they been applied to any of following domains, such as deep fake videos, or biological signals.

\begin{figure*}[h!]
\centering
  \includegraphics[width=0.9\linewidth]{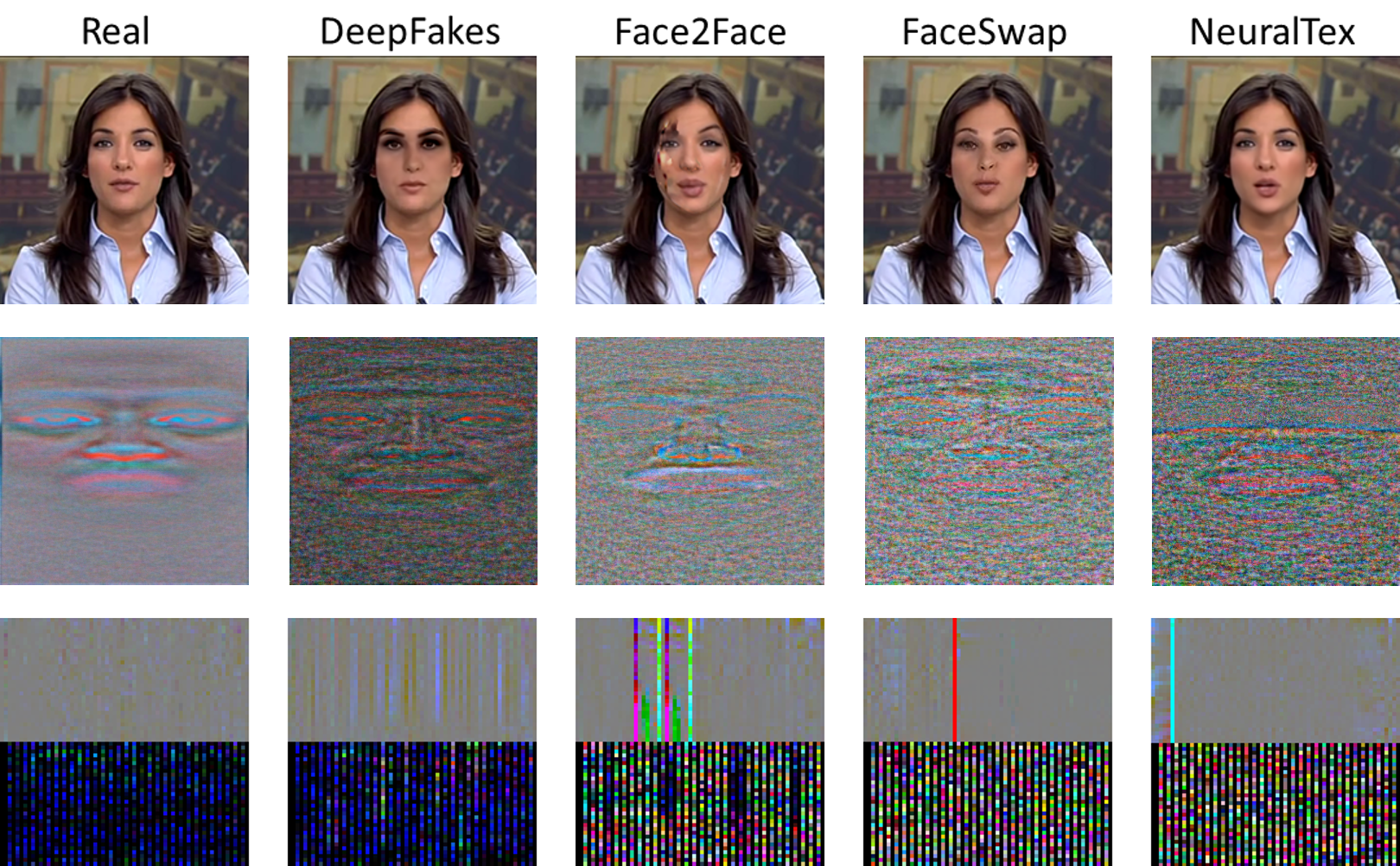}
 \caption{\textbf{PPG Cells}. Example frames per $\omega=64$ window (top), and their PPG cells (bottom) consisting of raw PPG and PPG PSD, of a real video (left) and its deep fakes per generative model (rest). Middle row represents an approximation to the accumulated residuals over all videos, which correlates with the colors in the PPG spectra.}
  \label{fig:ppg}
\end{figure*}
\subsection{Deep Fake Datasets}

With the increase of deep fake generation, the exigencies of detecting such doctored data become essential. While a large amount of such videos/images spread on internet or social media, it is highly demanded to have benchmark datasets specifically curated for research of deep fake detection. With respect to the data source generation, we categorize the existing datasets by two types: (1) datasets with single model generation and (2) datasets using multiple generative sources. Although there exist several synthetic face image datasets~\cite{Karras_2019_CVPR, GeneratedFacesDataset, neves2019ganprintr}, here we focus on video datasets due to the absence of biological signals in single images. 

The majority of existing deep fake video datasets contain videos created by single, easy-access, and popular generative sources. For instances, UADFV~\cite{8683164} dataset contains 48 real and 48 fake videos generated by FakeAPP~\cite{FakeApp}. DeepfakeTIMIT~\cite{vidTimid} dataset has 650 deep fake videos generated using faceswap-GAN~\cite{FaceSwap-Gan} where vidtimit~\cite{10.1007/978-3-642-01793-3_21} videos are used as originals. FaceForensics~\cite{ff} dataset congregates 1,004 videos from the internet with their deep fake versions created by Face2Face~\cite{f2f}, resulting in 2,008 videos. Celeb-DF~\cite{Celeb_DF_cvpr20} dataset collects 590 real videos of famous actors, with 5,639 deep fake versions generated by an improved synthesis process~\cite{Celeb_DF_cvpr20}. 

A typical dataset generated by multiple generative methods is the commonly used FaceForensics++~\cite{FF++} (FF) dataset, which includes 1,000 real videos and 4,000 fake videos, generated by four generative models \--- FaceSwap~\cite{FaceSwap}, Face2Face~\cite{f2f}, Deepfakes~\cite{DeepFakes}, and Neural Textures~\cite{10.1145/3306346.3323035}. Recently, an in-the-wild deep fake dataset was created by Ciftci et al.  \cite{FakeCatcher}, in which 140 videos are collected online, and half of them are fake. However, source models of those fake videos are unknown, thus posing a big challenge for in-the-wild deep fake source detection.  

\section{PPG Cells}

Biological signals have been proven as an authenticity indicator for real videos, which have been used as a distinguishable biomarker for deep fake detection~\cite{FakeCatcher}. As we know, a synthetic person shown in a fake video does not exhibit a similar pattern of heart beat as the one shown in a real video does~\cite{FakeCatcher}. \textit{Our key finding emerges from the fact that we can interpret these biological signals as fake heart beats that contain a signature transformation of the residuals per model.} Thus, it gives rise to a new exploration of these biological signals for not only determining the authenticity of a video, but also classifying its source model that generates the video. Our proposed system for detection of both deep fakes and their sources is outlined in Figure~\ref{fig:teaser}. 

In order to capture the characteristics of biological signals consistently, we define a novel spatiotemporal block, called the PPG cell. The PPG cells combine several raw PPG signals and their power spectra, extracted from a fixed window. The generation of PPG cells starts with finding the face in every frame using a face detector \cite{openface}. In case the window contains multiple faces, we process signals individually and aggregate the results in the final step.

The second step is to extract regions of interests (ROI) from the detected faces that have as much stable PPG signals as possible (Figure~\ref{fig:teaser}(d)). Biological signals are sensitive to facial movements, illumination variations, and facial occlusions. In order to extract these areas robustly, we use the face region between eye and mouth regions, maximizing the skin exposure. As the PPG signals from different face regions are correlated with each other \cite{FakeCatcher}, locating the ROIs and measuring their correlation become a crucial step to enhance the detection. 

The third step involves aligning these nonlinear ROIs to a rectangular image. We employ Delaunay triangulation~\cite{Fortune:1997:VDD:285869.285891}, followed by a nonlinear affine transformation per triangle to transform each triangle into the rectified image.

In the fourth step, we divide each image into 32 equal-size squares and calculate the raw Chrom-PPG signal per square in a fixed window with the size of $\omega$ frames, without interruptions in face detection (Figure~\ref{fig:teaser}(e)). Then, we calculate the Chrom-PPG in the rectified image~\cite{7867752} since it produces more reliable PPG signals~\cite{7565547,7867752}. For each window we now have $\omega$ times 32 raw PPG values. We reorganize these into a matrix of 32 rows and $\omega$ columns, forming the base of the PPG cells as shown in Figure~\ref{fig:teaser}(f) and Figure~\ref{fig:ppg} top half of bottom rows. Note that the bright columns correspond to significant motion or illumination changes where the PPG signal deviates abruptly.

The final step adds the information from the frequency domain to the PPG cells. We calculate the power spectral density of each raw PPG value in the window and scale it to $\omega$ size. We concatenate the power spectra to the bottom to generate PPG cells with 64 rows and $\omega$ columns (Figure~\ref{fig:teaser}(f)). Figure~\ref{fig:ppg} bottom row shows example PPG cells of deep fakes generated from the same window, with an example frame from each window at the top row. To analyze the contribution of the spectral information, we conduct experiments on PPG cells both with and without this last step and compare their accuracies (Section \ref{Sec:AblationStudy}).

Having defined PPG cells, now we can demonstrate a claim for our main hypothesis: the projection of residuals of deep fake generators into the biological signal domain creates a unique pattern that can be utilized for source detection. As proposed by ~\cite{neves2019ganprintr}, GAN residuals can be approximated by consistent noise in fake images. We apply temporal non-local means denoising on the aligned face in one frame from each video in FF. We then accumulate and normalize the difference of original and denoised images, and subtract the noise of the real images from each corresponding fake residual to obtain the middle row in Figure~\ref{fig:ppg}, containing the "fingerprint" per generator. For the real class, we demonstrate the overall noise accumulation. The colors of PPG-PSD correspond to different frequencies in the spectra of these residuals, and some of these frequencies are actually visible in the residual accumulation images. Our main observation follows this correlation between the residuals and our PPG cells: residuals create unique variations in the ``deep fake heart beats'' per model.

\section{Model Architecture}

As introduced in the related work section, the state of the art fake detection approaches employ binary classification techniques. For this binary classification task, even shallow CNNs are demonstrated to be useful with the addition of biological signals~\cite{FakeCatcher} when compared to complex network architectures without biological signals. However, as we take one step further from these approaches and introduce multiple classes for source detection, we need a more complex feature space segmentation, thus we put more emphasis on the deep learning model architecture. We formulate this as a multi-label classification task with equally probable classes of different generative sources and real videos.

Our learning setting is built on the FaceForensics++ (FF) dataset with a 70\%-vs-30\% split, where we generate PPG cells with a window size of $\omega=128$. FF dataset contains 4 different generative models, and we add real videos as the fifth class. Using a simple CNN with 3 VGG~\cite{simonyan2014deep} blocks, we achieve 68.45\% accuracy for PPG cell classification on 5 classes in the FF dataset, showing the need for a higher capacity model. Extending with another VGG block results in 75.49\%, confirming our intuition. Both to follow this intuition and also to keep our implementation simple, we experiment with VGG16~\cite{simonyan2014deep}, VGG19~\cite{simonyan2014deep}, InceptionV3~\cite{Szegedy_2016_CVPR}, Xception~\cite{Chollet_2017_CVPR}, ResNet50~\cite{he2015deep}, DenseNet201~\cite{DBLP:journals/corr/HuangLW16a}, and MobileNet~\cite{DBLP:journals/corr/HowardZCKWWAA17}, training for 100 epochs, with $\omega=128$, using the same 70\%-vs-30\% split. Table~\ref{tab:NN_PPG_ACC} lists the results of PPG cell classification on the test set, where VGG19 achieves the highest accuracy for differentiating the 4 different generative models and real videos of FF (Figure~\ref{fig:teaser}(f)). Complex networks like DenseNet and MobileNet overfit, reaching a very high training accuracy, but failing on the test set.

\begin{table}[h]
\centering    
\begin{tabular}{c|c|c}
Backbone  & FD Accuracy & SD Accuracy
\\\hline
ResNet50     &19.31\% &52.23\%\\
MobileNet    &27.26\% &33.16\%\\ 
Inception    &52.81\% &58.60\%\\
DenseNet201  &30.82\%  &37.04\%\\
Xception     &70.72\% &68.54\%\\
VGG16        &71.83\% &76.94\%\\
VGG19        &\textbf{76.15\%} &\textbf{81.06\%}\\
\end{tabular}
\caption{\textbf{PPG Cell Classification Accuracy.} Overall accuracies with different models for fake detection (FD) as binary classification and source detection (SD) as multi-class classification with $\omega=128$ on FF dataset.}
\label{tab:NN_PPG_ACC}
\end{table} 

\section{Video Classification}
\label{Sec:VideoClassification}

Even though our $\omega$-frames PPG cells can act as mini videos, a full video consists of several windows of PPG cells, depending on its length. Therefore we need to aggregate per-cell predictions into per-video predictions. Instead of brute force majority voting, we exploit the prediction confidences and employ log of odds to output the final video accuracies (Figure~\ref{fig:teaser}(f)). We document different voting schemes for this process in Table~\ref{tab:AGG_Methods}, where we compare majority voting, highest average probabilities, two highest average probabilities, and average of log odds on our cell prediction results by VGG19 using $\omega=128$. Average logits increase the video source classification accuracy to 84.93\%, 0.46\% higher than majority voting, as it is more robust against outliers by utilizing all predictions for all classes of PPG cells for a given video. We would like to conclude this section by noting that the longer the video is, the more PPG cells we have, and the stronger predictions our system will make, based on this aggregation process.

\begin{table}[h]
\centering    
\begin{tabular}{c|c}
Aggregation  & Video SD Accuracy
\\\hline
majority voting      &84.47\%\\
$\langle \rho \rangle$, where $\rho > 50\%$ &83.53\%\\
$\rho_{max}$ &83.60\%\\
$\langle \{ \rho_{max_1}, \rho_{max_2} \} \rangle$ &83.19\%\\
$\langle log{\frac{\rho}{1-\rho}} \rangle$ &\textbf{84.93}\%\\
\end{tabular}
\caption{\textbf{Prediction Aggregation from PPG Cell Classification.} Video source detection accuracies based on different voting schemes for the prediction probabilities ($\rho$). $\langle . \rangle$ denotes the mean.}
\label{tab:AGG_Methods}
\end{table}

\section{Results}

Our system is implemented in python utilizing OpenFace~\cite{openface} library for face detection, OpenCV~\cite{opencv_library} for image processing, and Keras~\cite{chollet2015keras} for neural network implementations. Most of the training and testing is performed on a desktop with a single NVIDIA GTX 1060 GPU, with tractable training times. The most computationally expensive part of the system is the extraction of PPG cells from large datasets, which is a one time process per video. In this section we document our analysis, results, and some ablation studies. Unless otherwise noted, we set our testbed as the FF dataset with the same 70\%-vs-30\% split \--- 700 real videos and 4*700 deep fakes for training, and 300 real videos and 4*300 deep fakes for testing. 

\subsection{Source Classification Accuracy}
\begin{figure}[h!]
\centering
  \includegraphics[width=1\linewidth]{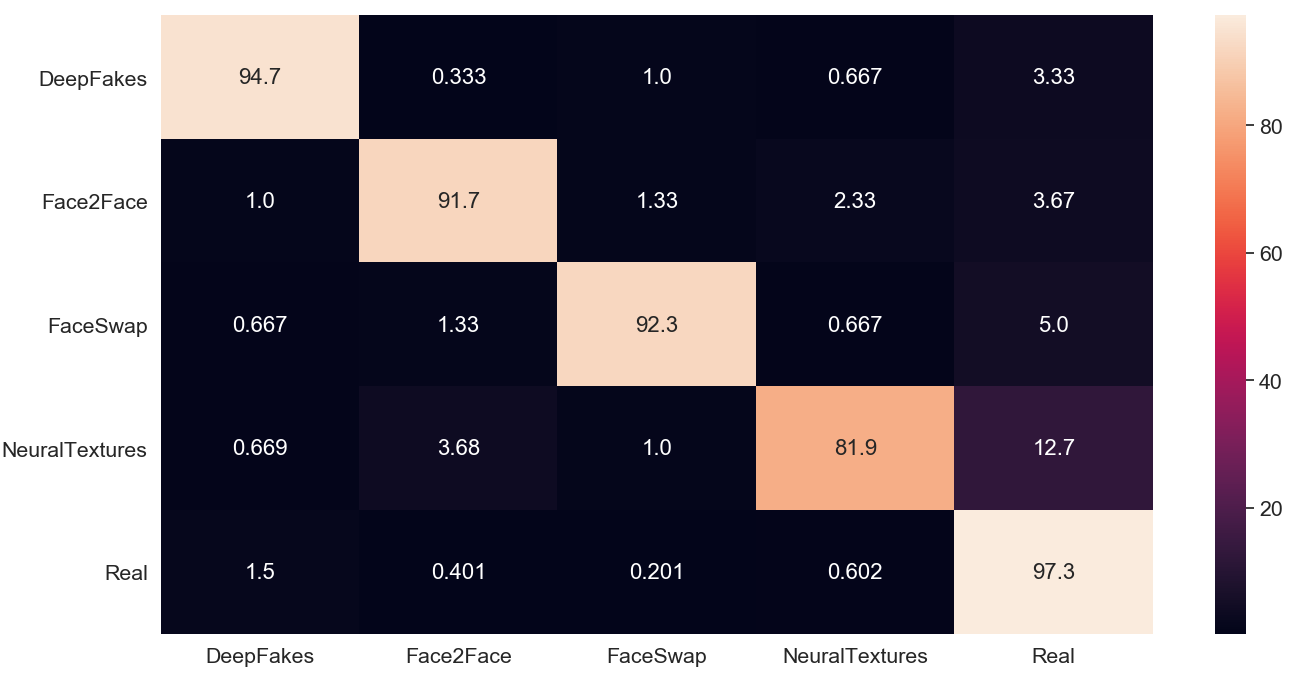}
 \caption{\textbf{Confusion Matrix for Class Accuracies}. Video source detection accuracies per 4 generative models and real videos with $\omega=64$ on the FF test set, with an average of 93.39\% accuracy.}
  \label{fig:conf}
\end{figure}

To better evaluate our video source classification, we analyze how uniquely each generative model is detected using the biological signals as a modulator for residuals. This analysis supports our claim of different generative models having signature patterns projected to the biological signal space. As per Figure~\ref{fig:conf}, our approach correctly detects real videos with 97.3\%, and generative models with at least 81.9\% accuracy for five classes (1 real and 4 fakes) of FF. 

\subsection{Ablation Study}
\label{Sec:AblationStudy}
In this section, first we train and test on different setups, namely (i) without real videos in the training set, (ii) without the power spectrum in PPG cells, (iii) without biological signals, and (iv) using full frames instead of face ROIs, where $\omega=64$ and FF dataset split are set as constants. In the second part, we analyze the effect of $\omega$.

\subsubsection{Different Setups}

Comparing the first two columns in Table~\ref{tab:abla}, overall accuracy very slightly increases, which may be expected when there are less number of classes. Since there exist several binary deep fake detectors, this test ensures that our method can be used as a secondary step to detect the source, after a video is determined to be fake.

\begin{table}[h]
\centering    
\begin{tabular}{c|c|c|c|c|c}
Source      & All  & -real  & -PSD & -PPG & Full
\\\hline
DeepFakes   & 94.7 & 94.66  &94.66 &93.00 & 57.33\\
Face2Face   & 91.7 & 91.66  &94.66 &87.33 & 37.33\\
FaceSwap    & 92.3 & 94.00  &94.66 &92.66 & 45.66\\
NeuralTex   & 81.9 & 93.07  &85.95 &83.61 & 41.33\\
Real        & 97.3 & NA     &89.66 &87.20 & 51.00\\
\hline
Total       &93.39 & 93.57  &92.11 &88.77 & 46.53\\
\end{tabular}
\caption{\textbf{Ablation Study.} Video source detection accuracies without reals, without PSD part of PPG cells, without biological signals, and on full frames (not only faces).}
\label{tab:abla}
\end{table} 
Comparing the first column to the third column, overall increase is only 1.28\% in accuracy. However detection of real videos has an increase of 7.64\%, which confirms the main contribution of the power spectrum: the spatiotemporal correlation of biological signals in real videos is not preserved in deep fakes, so it is useful in authenticity detection. The last two columns re-justify the contributions of \cite{FakeCatcher} that (i) biological signals are a crucial factor in fake detection, and (ii) training on faces instead of full frames improves the accuracy.

\subsubsection{Window Length}
The duration from which to extract PPG signals plays an important role in the stability and representative power of the PPG cells. Short windows may miss PPG frequencies and long windows may include too much noise to overshadow the actual signal. We test our method with different window sizes of $\omega=\{64, 128, 256, 512\}$ frames to balance these ends, on the same setup discussed before (Table~\ref{tab:SequenceSize}). With an optimum of $\omega=64$, as we increase the window length, PSNR decrease, and the accuracy drops.

\begin{table}[h]
\centering    
\begin{tabular}{c|c|c|c|c}
Source & $\omega=64$ & $\omega=128$ & $\omega=256$ & $\omega=512$ 
\\\hline
DeepFakes      &94.66 &93.62&93.26&88.99\\
Face2Face      &91.66&87.62&85.23&69.29\\
FaceSwap       &92.33&90.96&83.94&78.26\\
NeuralTex &81.93&69.23&84.56&31.11\\
Real           &97.29&83.27&82.88&78.89\\
\hline
Total            &\textbf{93.39}&84.93&85.97&73.75\\

\end{tabular}
\caption{\textbf{Effect of Window Length.} Video source detection with varying $\omega$ frame windows.}
\label{tab:SequenceSize}
\end{table}

\subsection{Extending with New Models}
Although we have detected four generative models, it is still a challenging task as new deep fake sources emerge rapidly. To justify that our approach can extend to new models, we combine our FF setup with the single generator dataset CelebDF\cite{Celeb_DF_cvpr20} and repeat the analysis. We randomly select 1,000 fake videos from CelebDF and create a sixth class for their generative model. Our approach achieves 93.69\% overall accuracy with 92.17\% accuracy on CelebDF, concluding that we can adapt to new models (Table~\ref{tab:CelebDB1000Fake}). We emphasize that we do not need real counterparts, we only train on fake samples.

\begin{table}[h]
\centering    
\begin{tabular}{c|c}
Source & Video SD Accuracy
\\\hline
CelebDF        &92.17\%\\
DeepFakes       &94.66\%\\
Face2Face       &91.66\%\\
FaceSwap        &92.66\%\\
NeuralTex  &86.62\%\\
Real            &96.89\%\\
\hline
Total           &93.69\%\\
\end{tabular}
\caption{\textbf{New Model Extension.} Video source detection accuracies with 1000 fakes from CelebDF added to FF, as a new class.}
\label{tab:CelebDB1000Fake}
\end{table}

This experiment also amplifies our motivation of detecting generative models from their residuals only. In contrast to other source detection methods utilizing the generator architecture or last layers for residual classification, we easily extend to new models without the need for the model specification or the real counterparts of the fake samples.


\subsection{Comparison}

To our knowledge, this paper leads the deep fake detection research towards source detection, utilizing biological signals to classify the residuals of generative models. Some image-based fake detection approaches have been proposed, however, there is no previous work that classifies deep fakes videos using biological signals that enables us to perform a one-to-one comparison with. Thus, we perform experiments with existing approaches on the aforementioned face ROIs,  such as (i) the same architecture on video frames (without biological signals), (ii) the same architecture on face ROIs, and (iii) frame-based classification approaches on face ROIs. 

Table~\ref{tab:Comparison} lists the accuracies of different models on the test set. The first row uses the same backbone (VGG19), but only on frames, without biological signals. In order to keep the training time tractable, we utilize every 20$^{th}$ frame. This can be thought as a baseline for frame-based detection. The second and third blocks are trained and tested on the same dataset, but on segmented and aligned face images. As generators are only swapping or modifying the faces, this approach both makes the training more efficient and improves the accuracies significantly. In this case, our method outperforms even the most complex network, Xception~\cite{Chollet_2017_CVPR}, with more than 10\% accuracy.

\begin{table}[h]
\centering    
\begin{tabular}{c|c}
Models      &Video SD Accuracy
\\\hline
VGG19 (frames)  & 46.53\%\\
\hline
VGG19 (faces)   &76.67\%\\
\hline
ResNet50        &63.25\%\\
ResNet152       &68.92\%\\
Inception       &79.37\%\\
DenseNet201     &81.65\%\\
Xception        &83.50\%\\
\hline
Ours            &\textbf{93.69\%}\\
\end{tabular}
\caption{\textbf{Comparison.} Video source detection accuracies on FF dataset, of several models with frame and face based training.}
\label{tab:Comparison}
\end{table} 

It is worth noting that our approach is advantageous in computational efficiency. As compared to the frame-based detection approach with the same architecture, which takes 29 hours 24 minutes and 43 seconds to train only one epoch with 1.8 million frames in FF on a single GPU, our approach takes only 2 hours and 35 minutes for training the system for 100 epochs. Such a computational efficiency in training and testing with large datasets makes our approach much more feasible to many application fields without demanding high-end computation powers.


\subsection{Unseen Generators}

Previously, we discussed that removal of real class improves the accuracy of finding the distinct residuals of the generative models. This emerges from the fact that PPG signals are affected not only by the generative model residual, but also environmental effects such as lighting, facial movement, and occlusion. As such random artifacts cannot create a pattern, all of those PPG deviations are classified as real, as real is the ``chaotic'' state without an exact signature. In order to test this hypothesis, we congregate a new test dataset from UADFV~\cite{8683164}, FaceForensics~\cite{ff}, Deep Fakes Dataset~\cite{FakeCatcher}, and CelebDB~\cite{Celeb_DF_cvpr20} where we gather 48 real and fake video pairs from each (equal to the smallest of these datasets). On this new collection of 384 videos, we run our previous best model trained on FF with $\omega=64$, with and without the real class. 

\begin{figure}[h!]
\centering
  \includegraphics[width=1\linewidth]{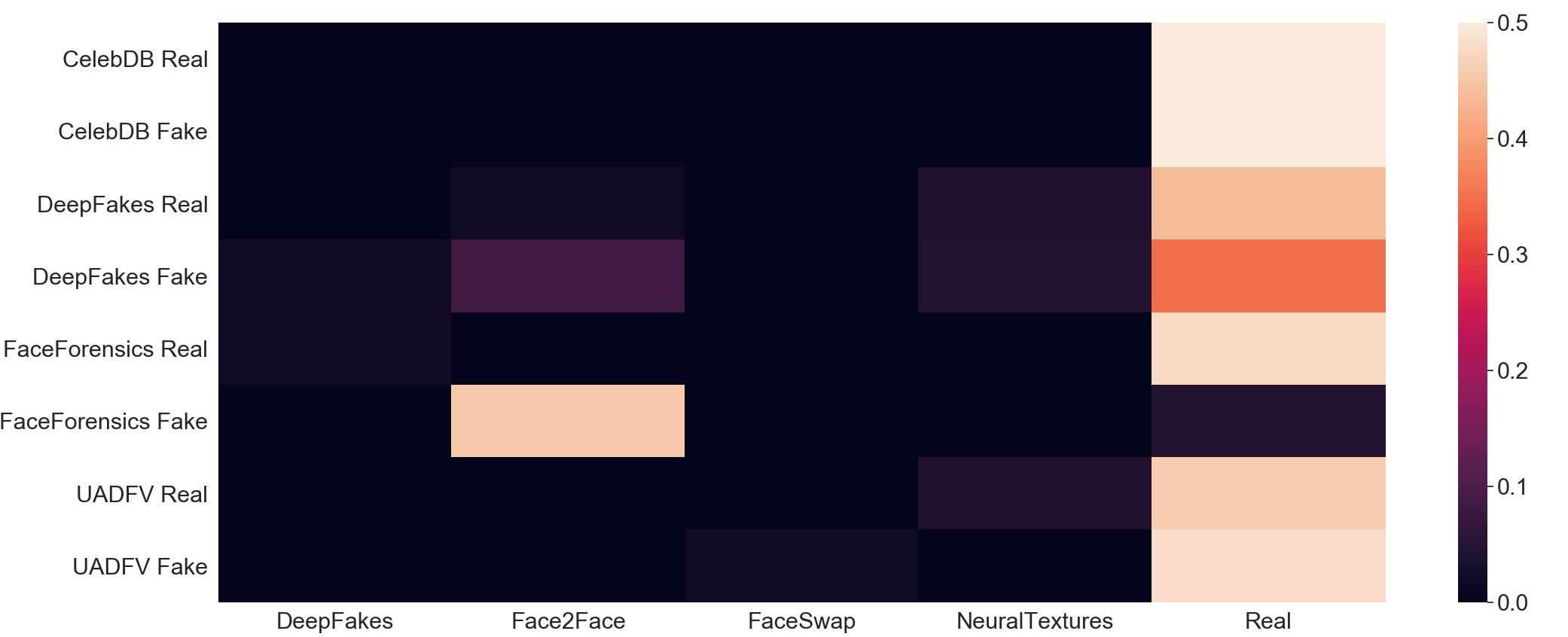}
 \caption{\textbf{Unseen Datasets}. Fake and real classification of 96*4 videos from single and multi source unseen datasets.}
  \label{fig:bigconf}
\end{figure}

In addition to the confusion matrix in Figure~\ref{fig:conf}, we depict these new classifications in Figure~\ref{fig:bigconf}. To begin with, true positives for reals are 100\%, 93.61\%, 97.82\%, and 95.83\% according to column 5, rows 1, 3, 5, and 7 respectively. Confirming our hypothesis, UADFV and CelebDF classifications are expected to tend towards the real class (col 5, rows 2\&4), because the model does not recognize their signature yet. FaceForensics is expected to be classified as Face2Face~\cite{f2f} (col 2, row 6), as its generative model is within FF.  Deep Fake Datasets should have a variety of classification results (row 4) as it contains in-the-wild videos with unknown generators. 

\section{Conclusion}

In this paper, we present a deep fake source detection technique via interpreting residuals with biological signals. To our knowledge this is the first method to apply biological signals to the task of deep fake source detection. In addition we experimentally validate our method through various ablation studies. In our experiments we achieve 93.39\% accuracy on FaceForensics++~\cite{FF++} dataset on source detection from four deep fake generators and real videos. Moreover, we demonstrate the adaptability of our approach to new generative models, keeping the accuracy unchanged.

Following the study in biological signal analysis on deep fake videos, the ground truth PPG data along with real and fake videos can enable a novel direction in research on deep fake analysis and detection. In the next stage of the work, we plan to create a new dataset with ground truth PPG, with certain source variations as well as distribution variations. 

It is worth noting that our work looks for generator signatures in deep fakes, while the existing work reported by Ciftci et al. \cite{FakeCatcher} looks for signatures in real videos. Theoretically, a holistic system combining these two perspectives can be developed with a jointly trained model for detecting signatures on both authentic and fake videos. We pose this idea as our immediate future work.

\section*{Acknowledgment}
This work is supported in part by the National Science Foundation under grant CNS-1629898. 

{\small
\bibliographystyle{ieee}
\bibliography{submission_example}
}

\end{document}